\documentclass[10pt,twocolumn,letterpaper]{article}

\usepackage{cvpr}
\usepackage{times}
\usepackage{epsfig}
\usepackage{graphicx}
\usepackage{amsmath}
\usepackage{amssymb}
\usepackage{bm}
\usepackage{xcolor}
\usepackage{booktabs}
\usepackage{soul}
\usepackage{subfig}
\usepackage[font={small}]{caption}

\usepackage[pagebackref=true,breaklinks=true,letterpaper=true,colorlinks,bookmarks=false]{hyperref}
\usepackage[british,UKenglish,USenglish,american]{babel}

\cvprfinalcopy 

\def\cvprPaperID{***} 

\ifcvprfinal\fi
\begin{document}

\title{FaceShifter: Towards High Fidelity And Occlusion Aware Face Swapping}
\author{Lingzhi Li$^{1*}$ \qquad Jianmin Bao$^{2\dag}$ \qquad Hao Yang$^{2}$ \qquad Dong Chen$^{2}$ \qquad Fang Wen$^{2}$ \qquad \vspace{1pt}\\
$^{1}$Peking University  \qquad $^{2}$Microsoft Research\qquad\qquad\\
\hspace{0.1in}{\tt\small lilingzhi@pku.edu.cn} \qquad  {\tt\small \{jianbao,haya,doch,fangwen\}@microsoft.com} \\
}

\twocolumn[{%
\renewcommand\twocolumn[1][]{#1}%
\vspace{-1em}
\maketitle
\vspace{-1em}
\begin{center}
\centering

\includegraphics[width=1.0\linewidth]{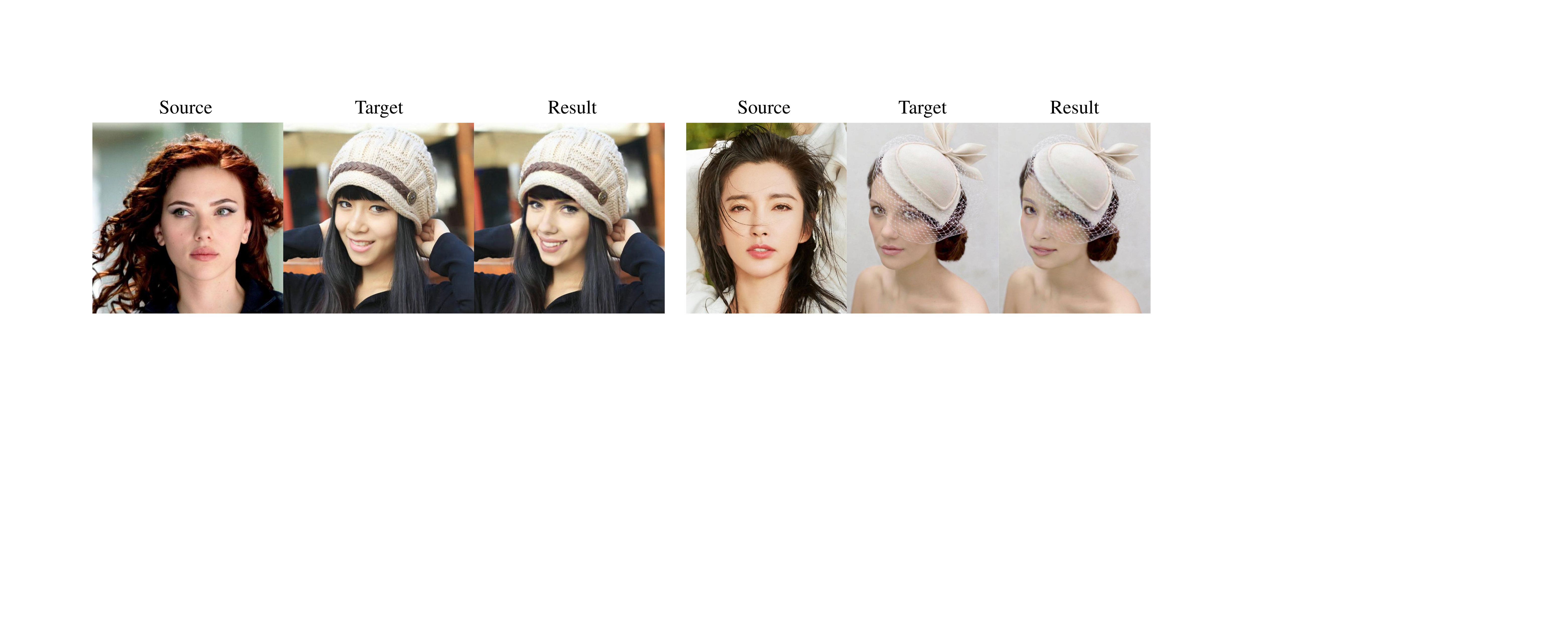}
\captionof{figure}{The face in the source image is taken to replace the face in the target image. Results of FaceShifter appear in the right.
}
\label{fig:teaser}
\end{center}%
}]

\maketitle

\begin{abstract}
In this work, we propose a novel two-stage framework, called FaceShifter, for high fidelity and occlusion aware face swapping. 
Unlike many existing face swapping works that leverage only limited information from the target image when synthesizing the swapped face, 
our framework, in its first stage, generates the swapped face in high-fidelity by exploiting and integrating the target attributes thoroughly and adaptively. 
We propose a novel attributes encoder for extracting multi-level target face attributes, and a new generator with carefully designed Adaptive Attentional Denormalization (AAD) layers to adaptively integrate the identity and the attributes for face synthesis. 
To address the challenging facial occlusions, we append a second stage consisting of a novel Heuristic Error Acknowledging Refinement Network (HEAR-Net). It is trained to recover anomaly regions in a self-supervised way without any manual annotations. 
Extensive experiments on wild faces demonstrate that our face swapping results are not only considerably more perceptually appealing, but also better identity preserving in comparison to other state-of-the-art methods. For generated dataset and more detail please refer to our project webpage \url{https://lingzhili.com/FaceShifterPage/} {\let\thefootnote\relax\footnotetext{$^{*}$ Work done during an internship at Microsoft Research Asia}}
{\let\thefootnote\relax\footnotetext{$^{\dag}$ Corresponding Author}}
\end{abstract}

\section{Introduction}

Face swapping is the replacement of the identity of a person in the target image with that of another person in the source image, while preserving attributes \eg head pose, facial expression, lighting, background \etc. Face swapping has attracted great interest in vision and graphics community, because of its potential wide applications in movie composition, computer games, and privacy protection~\cite{ross2010visual}.

The main difficulties in face swapping are how to extract and adaptively recombine identity and attributes of two images. Early replacement-based works~\cite{bitouk2008face,wang2008facial} simply replace the pixels of inner face region. Thus, they are sensitive to the variations in posture and perspective. 3D-based works~\cite{blanz2004exchanging,cheng20093d,lin2012face,nirkin2018face} used a 3D model to deal with the posture or perspective difference. However the accuracy and robustness of 3D reconstruction of faces are all unsatisfactory. Recently, GAN-based works~\cite{korshunova2017fast,natsume2018fsnet,natsume2018rsgan,nirkin2019fsgan, bao2017cvae} have illustrated impressive results. But it remains challenging to synthesize both realistic and high-fidelity results. 

In this work, we focus on improving the fidelity of face swapping. 
In order to make the results more perceptually appealing, it is important that the synthesized swapped face not only shares the pose and expression of the target face, but also can be seamlessly fitted into the target image without inconsistency: the rendering of the swapped face should be faithful to the lighting (\eg direction, intensity, color) of the target scene, the pixel resolution of the swapped face should also be consistent with the target image resolution. 
Neither of these can be well handled by a simple alpha or Poisson blending.
Instead, we need a \emph{thorough and adaptive integration of target image attributes during the synthesis of the swapped face}, so that the attributes from the target image, including scene lighting or image resolution, can help make the swapped face more realistic.

\begin{figure}[t]
\centering
\includegraphics[width=\linewidth]{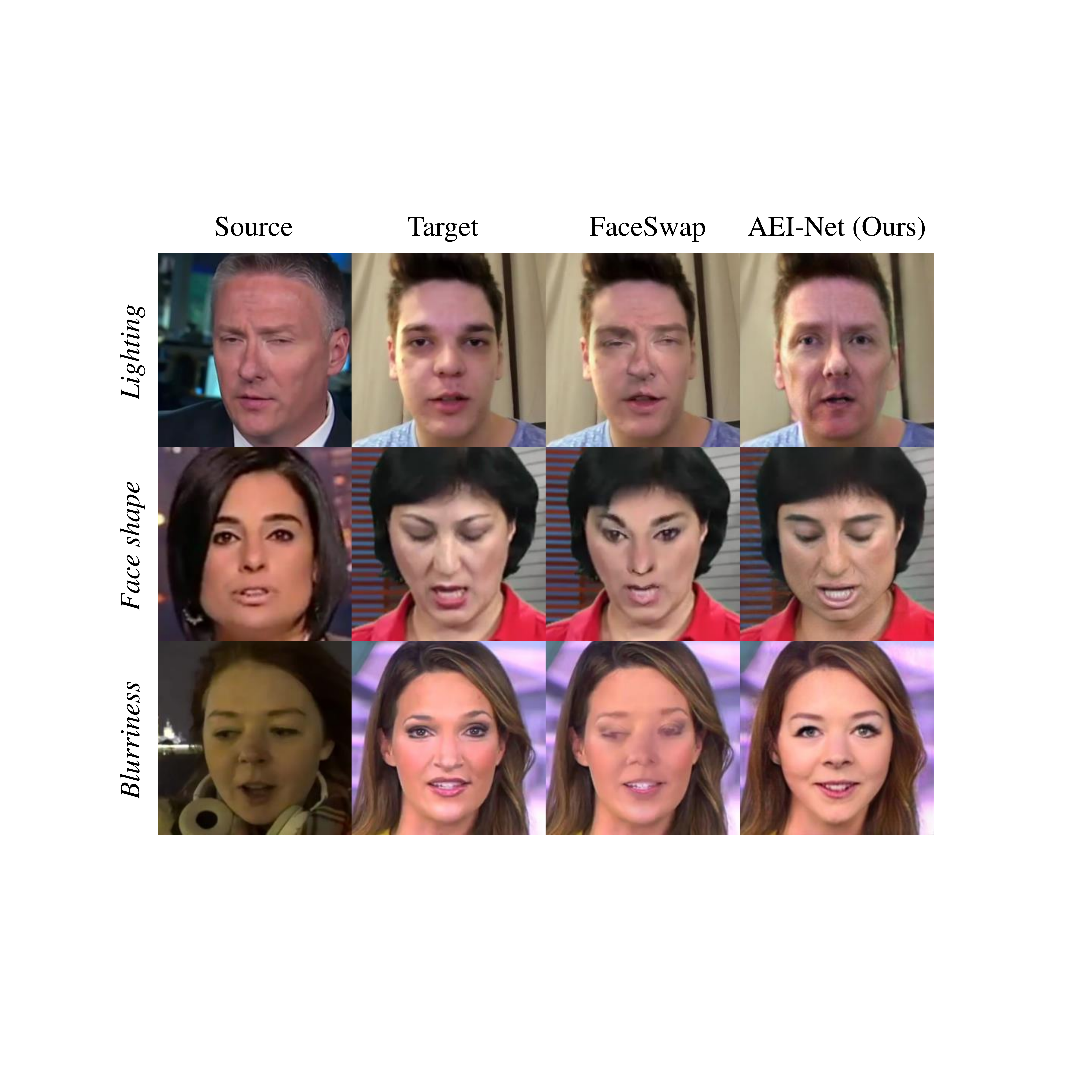}
\caption{
Failure cases of a previous method on FaceForensics++ \cite{rossler2019faceforensics++} dataset. 
From left to right we show the input source images, the input target images, the results of FaceSwap \cite{faceswap}, and the results of our method.
FaceSwap follows the strategy that, first synthesizes the inner face region, then blends it into the target face. 
Such strategy causes artifacts, such as the defective lighting effect on the nose (row 1), failing to preserve the face shape of the source identity (row 2) and the mismatched image resolutions (row 3).  While our method addresses all these issues. 
}
\label{fig:existing_issues}
\vspace{-1em}
\end{figure}

However, previous face swapping methods either neglect the requirement of this integration, or lack the ability to perform it in a thorough and adaptive way.
In specific, many previous methods use only pose and expression guidances from the target image to synthesize the swapped face, the face is then blended into the target image using masks of the target faces. This process is easy to cause artifacts, because:
1) Besides pose and expression, it leverages little knowledge about the target image when synthesizing the swapped face, which can hardly respect target attributes like the scene lightings or the image resolutions; 
2) Such a blending will discard all the peripheral area of the source face that locates outside the target face mask. Thus these methods cannot preserve the face shape of the source identity.
We show some typical failure cases in Figure \ref{fig:existing_issues}. 

In order to achieve high-fidelity face swapping results, in the first stage of our framework, we design a GAN-based network, named \emph{Adaptive Embedding Integration Network} (AEI-Net), for a thorough and adaptive integration of target attributes. We made two improvements to the network structure: 1) we propose a novel \emph{multi-level attributes encoder} for extracting target attributes in various spatial resolutions, instead of compressing it into a single vector as RSGAN~\cite{natsume2018rsgan} and IPGAN~\cite{Bao_ipgan}. 2) we present a novel generator with carefully designed \emph{Adaptive Attentional Denormalization} ({AAD}) layers which adaptively learns where to integrate the attributes or identity embeddings. Such an adaptive integration brings considerable improvements over the single level integration used by RSGAN~\cite{natsume2018rsgan}, FSNet~\cite{natsume2018fsnet} and IPGAN~\cite{Bao_ipgan}. With these two improvements, the proposed AEI-Net can solve the problem of inconsistent illumination and face shape, as shown in Figure~\ref{fig:existing_issues}.

Moreover, handling facial occlusions is always challenging in face swapping. Unlike Nirkin \etal~\cite{nirkin2019fsgan,nirkin2018face} that trains face segmentation to obtain occlusion-aware face masks, our method can learn to recover face anomaly regions in a self-supervised way without any manual annotations. 
We observe that when feeding the same face image as both the target and source into a well trained AEI-Net, the reconstructed face image deviates from the input in multiple areas, these deviations strongly hint the locations of face occlusions. Thus,
we propose a novel \emph{Heuristic Error Acknowledging Refinement Network} ({HEAR-Net}) to further refine the result under the guidance of such reconstruction errors. The proposed method is more general, thus it identifies more anomaly types, such as glasses, shadow and reflection effects, and other uncommon occlusions.

The proposed two-stage face swapping framework, FaceShifter, is subject agnostic. Once trained, the model can be applied to any new face pairs without requiring subject specific training as DeepFakes~\cite{deepfake} and Korshunova~\etal~\cite{korshunova2017fast}. Experiments demonstrate that our method achieves results considerably more realistic and more faithful to inputs than other state-of-the-art methods.

\section{Related Works}

Face swapping has a long history in vision and graphics researches. Early efforts \cite{bitouk2008face,wang2008facial} only swap faces with similar poses. Such a limitation is addressed by recent algorithms roughly divided in two categories: 3D-based approaches and GAN-based approaches.

\noindent\textbf{3D-Based Approaches}. Blanz \etal \cite{blanz2004exchanging} considers 3D transform between two faces with different poses, but requiring user interaction and not handling expressions. 
Thies \etal \cite{thies2015real} captures head actions from a RGB-D image using 3DMM, turning a static face into a controllable avatar. It is extended for RGB references in Face2Face \cite{thies2016face2face}. 
Olszewski \etal \cite{olszewski2017realistic} dynamically inferences 3D face textures for improved manipulation quality.
Kim \etal \cite{kim2018deep} separately models different videos using 3DMM to make the portraits controllable, while Nagano \etal \cite{nagano2018pagan} needs only one image to reenact the portrait within. 
Recently, Thies \etal \cite{thies2019deferred} adopt neural textures, which can better disentangle geometry in face reenactment.
However, when applied on face swapping, these methods hardly leverage target attributes like occlusions, lighting or photo styles. 
To preserve the target facial occlusions, Nirkin \etal \cite{nirkin2018face,nirkin2019fsgan} collected data to train an occlusion-aware face segmentation network in a supervised way, which helps predict a visible target face mask for blending in the swapped face. While our method find the occlusions in a self-supervised way without any manually annotations.

\noindent\textbf{GAN-Based Approaches}. 
In the GAN-based face swapping methods, Korshunova \etal \cite{korshunova2017fastfaceswap} swap faces like transfer styles. It separately models different source identities, such as a CageNet for Nicolas Cage, a SwiftNet for Taylor Swift.
The recently popular DeepFakes \cite{deepfake} is another example of such subject-aware face swapping: for each new input, a new model has to be trained on two video sequences, one for the source and one for the target. 

This limitation has been addressed by subject-agnostic face swapping researches: 
RSGAN \cite{natsume2018rsgan} learns to extract vectorized embeddings for face and hair regions separately, and recombines them to synthesize a swapped face.
FSNet \cite{natsume2018fsnet} represents the face region of source image as a vector, which is combined with a non-face target image to generate the swapped face. 
IPGAN \cite{Bao_ipgan} disentangles the identity and attributes of faces as vectors. By introducing supervisions directly from the source identity and the target image, IPGAN supports face swapping with better identity preservation. However, due to the information loss caused by the compressed representation, and the lack of more adaptive information integration, these three methods are incapable of generating high-quality face images. Recently, FSGAN \cite{nirkin2019fsgan} performs face reenactment and face swapping together. It follows a similar reenact and blend strategy with \cite{olszewski2017realistic,nagano2018pagan}. Although FSGAN utilizes an occlusion-aware face segmentation network for preserving target occlusions, it hardly respects target attributes like the lighting or image resolution, it can neither preserve the face shape of the source identity.

\section{Methods}

Our method requires two input images, \ie., a source image ${X}_s$ to provide identity and a target image ${X}_t$ to provide attributes, \eg., pose, expression, scene lighting and background. The swapped face image is generated through a two-stage framework, called FaceShifter. In the first stage, we use an \emph{Adaptive Embedding Integration Network} (AEI-Net) to generate a high fidelity face swapping result $\hat{Y}_{s,t}$ based on information integration. In the second stage, we use the \emph{Heuristic Error Acknowledging Network} (HEAR-Net) to handle the facial occlusions and refine the result, the final result is denoted by $Y_{s,t}$.

\subsection{Adaptive Embedding Integration Network}
\label{sec: stage_I}

\begin{figure*}[t]
\centering
 \includegraphics[width=1.0\linewidth]{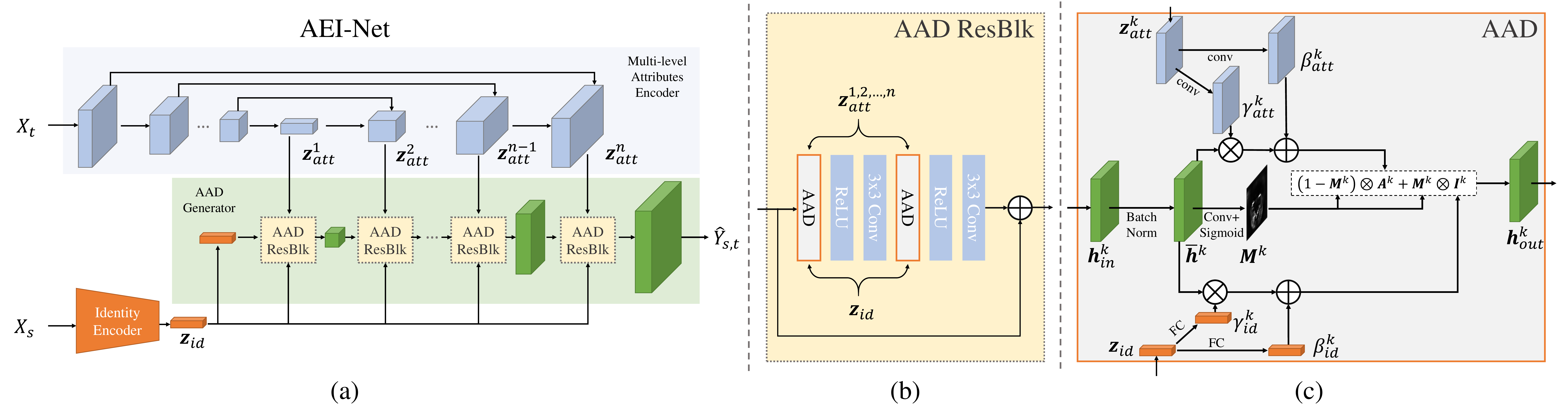}
 \footnotesize
    \caption{AEI-Net for the first stage. AEI-Net is composed of an Identity Encoder, a Multi-level Attributes Encoder, and an AAD-Generator. The AAD-Generator integrates informations of identity and attributes in multiple feature levels using cascaded AAD ResBlks, which is built on AAD layers.}
\label{fig:framework_stage1}
\vspace{-1em}
\end{figure*}

In the first stage, we aim to generate a high fidelity face image $\hat{Y}_{s,t}$, which should preserve the identity of the source ${X}_s$ and the attributes (\eg pose, expression, lighting, background) of the target ${X}_t$. To achieve this goal, our method consist of 3 components: i) the \emph{Identity Encoder} $\bm{z}_{id}(X_s)$, which extracts identity from the source image $X_s$; ii) the \emph{Multi-level Attributes Encoder} $\bm{z}_{att}(X_t)$, which extracts attributes of the target image $X_t$; iii) \emph{Adaptive Attentional Denormalization (AAD) Generator}, which generates swapped face image. Figure~\ref{fig:framework_stage1}(a) shows whole network structure.

\noindent\textbf{Identity Encoder}:
We use a pretrained state-of-the-art face recognition model \cite{deng2019arcface} as identity encoder. The identity embedding $\bm{z}_{id}(X_s)$ is defined to be the last feature vector generated before the final FC layer. 
We believe that by training on a large quantity of 2D face data, such a face recognition model can provide more representative identity embeddings than the 3D-based models like 3DMM \cite{blanz2004exchanging,blanz1999morphable}.

\noindent\textbf{Multi-level Attributes Encoder}:
Face attributes, such as pose, expression, lighting and background, require more spatial informations than identity. 
In order to preserve such details, we propose to represent the attributes embedding as multi-level feature maps,
instead of compressing it into a single vector as previous methods \cite{Bao_ipgan,natsume2018rsgan}.
In specific, we feed the target image $X_t$ into a U-Net-like structure. 
Then we define the attributes embedding as the feature maps generated from the U-Net decoder.
More formally, we define
\vspace{-0.5em}
\begin{equation}
\vspace{-0.1em}
\bm{z}_{att}(X_t) = \left\{\bm{z}^1_{att}(X_t), \bm{z}^2_{att}(X_t), \cdots \bm{z}^n_{att}(X_t)\right\},
\label{eqn:attribute_embedding}
\end{equation}
where $\bm{z}^{k}_{att}(X_t)$ represents the $k$-th level feature map from the U-Net decoder, $n$ is the number of feature levels. 

Our attributes embedding network does not require any attribute annotations, it extracts the attributes using self-supervised training: we require that the generated swapped face $\hat{Y}_{x_t}$ and the target image $X_t$ have the same attributes embedding. The loss function will be introduce in Equation~\ref{eqn:L_att}. In the experimental part (Section \ref{ssec:analysis}), we also study what the attributes embedding has learned.

\noindent\textbf{Adaptive Attentional Denormalization Generator}:
We then integrate such two embeddings $\bm{z}_{id}(X_s)$ and $\bm{z}_{att}(X_t)$ for generating a raw swapped face $\hat{Y}_{s,t}$.
Previous methods~\cite{Bao_ipgan,natsume2018rsgan} simply integrate them through feature concatenation. It will lead to relatively blurry results.
Instead, we propose a novel \emph{Adaptive Attentional Denormalization} (AAD) layer to accomplish this task in a more adaptive fashion.
Inspired by the mechanisms of SPADE \cite{park2019semantic} and AdaIN \cite{dumoulin2016learned, huang2017arbitrary}, the proposed AAD layers leverage denormalizations for feature integration in multiple feature levels.

As shown in Figure~\ref{fig:framework_stage1}(c), in the $k$-th feature level, 
let $\bm{h}_{in}^k$ denote the activation map that is fed into an AAD layer, which should be a 3D tensor of size $C^k \times H^k \times W^k$, with $C^k$ being the number of channels and $H^k \times W^k$ being the spatial dimensions.
Before integration, we perform batch normalization~\cite{bn} on $\bm{h}_{in}^k$: 
\vspace{-0.5em}
\begin{equation}
\vspace{-0.5em}
\label{eqn:IN}
\bar{\bm{h}}^k = \frac{\bm{h}_{in}^k - \bm{\mu}^k}{\bm{\sigma}^k}.
\end{equation}
Here $\bm{\mu}^k \in \mathbb{R}^{C^k}$ and $\bm{\sigma}^k \in \mathbb{R}^{C^k}$ are the means and standard deviations of the channel-wise activations within $\bm{h}_{in}^k$'s mini-batch. Then, we design 3 parallel branches from $\bar{\bm{h}}^k$ for 1) attributes integration, 2) identity integration, 3) adaptively attention mask.

For attributes embedding integration, let $\bm{z}_{att}^k$ be the attributes embedding on this feature level, which should be a 3D tensor of size $C^k_{att} \times H^k \times W^k$.
In order to integrate $\bm{z}_{att}^k$ into the activation, we compute an attribute activation $\bm{A}^k$ by denormalizing the normalized $\bar{\bm{h}}^k$ according to the attributes embedding, formulated as
\vspace{-0.5em}
\begin{equation}
\vspace{-0.5em}
\label{eqn:SPADE}
\bm{A}^k = \gamma^k_{att} \otimes \bar{\bm{h}}^k + \beta_{att}^k,
\end{equation}
where $\gamma^k_{att}$ and $\beta^k_{att}$ are two modulation parameters both convolved from $\bm{z}^k_{att}$. They share the same tensor dimensions with $\bar{\bm{h}}^k$. 
The computed $\gamma^k_{att}$ and $\beta^k_{att}$ are multiplied and added to $\bar{\bm{h}}^k$ element-wise.

For identity embedding integration, let $\bm{z}_{id}^k$ be the identity embedding, which should be a 1D vector of size $C_{id}$.
We also integrate $\bm{z}_{id}^k$ by computing an identity activation $\bm{I}^k$ in a similar way to integrating attributes. It is formulated as
\vspace{-0.5em}
\begin{equation}
\vspace{-0.5em}
\label{eqn:AdaIN}
\bm{I}^k = \gamma^k_{id} \otimes \bar{\bm{h}}^k + \beta_{id}^k,
\end{equation}
where $\gamma^k_{id} \in \mathbb{R}^{C^k}$ and $\beta^k_{id} \in \mathbb{R}^{C^k}$ are another two modulation parameters generated from $\bm{z}_{id}$ through FC layers.

One key design of the AAD layer is to adaptively adjust the effective regions of the identity embedding and the attributes embedding, so that they can participate in synthesizing different parts of the face. For example, the identity embedding should focus relatively more on synthesizing the face parts that are most discriminative for distinguishing identities, \eg eyes, mouth and face contour. 
Therefore, we adopt an attention mechanism into the AAD layer. 
Specifically, we generate an attentional mask $\bm{M}^k$ using $\bar{\bm{h}}^k$ through convolutions and a sigmoid operation. The values of $\bm{M}^k$ are between $0$ and $1$.

Finally, the output of this AAD layer $\bm{h}^k_{out}$ can be obtained as a element-wise combination of the two activations $\bm{A}^k$ and $\bm{I}^k$, weighted by the mask $\bm{M}^k$, as shown in Figure~\ref{fig:framework_stage1}(c). It is formulated as
\vspace{-0.5em}
\begin{equation}
\vspace{-0.5em}
\label{eqn:Mask}
 \bm{h}^k_{out} = (1 - \bm{M}^k) \otimes \bm{A}^k + \bm{M}^k \otimes \bm{I}^k.
\end{equation}

The AAD-Generator is then built with multiple AAD layers.
As illustrated in Figure \ref{fig:framework_stage1}(a), after extracting the identity embedding $\bm{z}_{id}$ from source $X_s$, and the attributes embedding $\bm{z}_{att}$ from target $X_t$, we cascade AAD Residual Blocks (AAD ResBlks) to generate the swapped face $\hat{Y}_{s,t}$, the structure of the AAD ResBlks is shown in Figure~\ref{fig:framework_stage1}(b). For the AAD ResBlk on the $k$-th feature level, it first takes the up-sampled activation from the previous level as input, then integrates this input with $\bm{z}_{id}$ and $\bm{z}^k_{att}$. The final output image $\hat{Y}_{s,t}$ is convolved from the last activation.

\noindent\textbf{Training Losses}
We utilize adversarial training for AEI-Net.
Let $\mathcal{L}_{adv}$ be the adversarial loss for making $\hat{Y}_{s,t}$ realistic. It is implemented as a multi-scale discriminator~\cite{park2019semantic} on the downsampled output images.
In addition, an identity preservation loss is used to preserve the identity of the source. It is formulated as
\vspace{-0.5em}
\begin{equation}
\vspace{-0.5em}
\label{eqn:L_id}
\mathcal{L}_{id} =  1 - cos(\bm{z}_{id}(\hat{Y}_{s, t}), \bm{z}_{id}(X_s) ),
\end{equation}
where $cos(\cdot,\cdot)$ represents the cosine similarity of two vectors. 
We also define the attributes preservation loss as $\mathcal{L}$-2 distances between the multi-level attributes embeddings from $X_t$ and $\hat{Y}_{s,t}$.
It is formulated as
\vspace{-0.5em}
\begin{equation}
\vspace{-0.5em}
\label{eqn:L_att}
\mathcal{L}_{att} = \frac{1}{2} \sum_{k=1}^n \left\|\bm{z}_{att}^k(\hat{Y}_{s,t}) - \bm{z}_{att}^k(X_t)\right\|_2^2.
\end{equation}
When the source and target images are the same in a training sample, we define a reconstruction loss as pixel level $\mathcal{L}$-2 distances between the target image $X_t$ and $\hat{Y}_{s,t}$
\vspace{-0.5em}
\begin{equation}
\vspace{-0.5em}
\label{eqn:L_rec1}
\mathcal{L}_{rec} = 
\begin{cases}
\frac{1}{2} \left\|\hat{Y}_{s,t} - X_t\right\|_2^2 & \textit{if}~X_t = X_s \\
0 & \textit{otherwise}
\end{cases}.
\end{equation}
The AEI-Net is finally trained with a weighted sum of above losses as
\vspace{-0.5em}
\begin{equation}
\vspace{-0.5em}
\label{eqn:L_stage1}
\mathcal{L}_\texttt{AEI-Net} = \mathcal{L}_{adv} + \lambda_{att} \mathcal{L}_{att} + \lambda_{id} \mathcal{L}_{id}  + \lambda_{rec} \mathcal{L}_{rec},
\end{equation}
with $\lambda_{att}=\lambda_{rec}=10, \lambda_{id}=5$.
The trainable modules of AEI-Net include the Multi-level Attributes Encoder and the ADD-Generator.

\subsection{Heuristic Error Acknowledging Refinement Network}
\label{sec: error_refine}

\begin{figure}[t]
\centering
 \includegraphics[width=\linewidth]{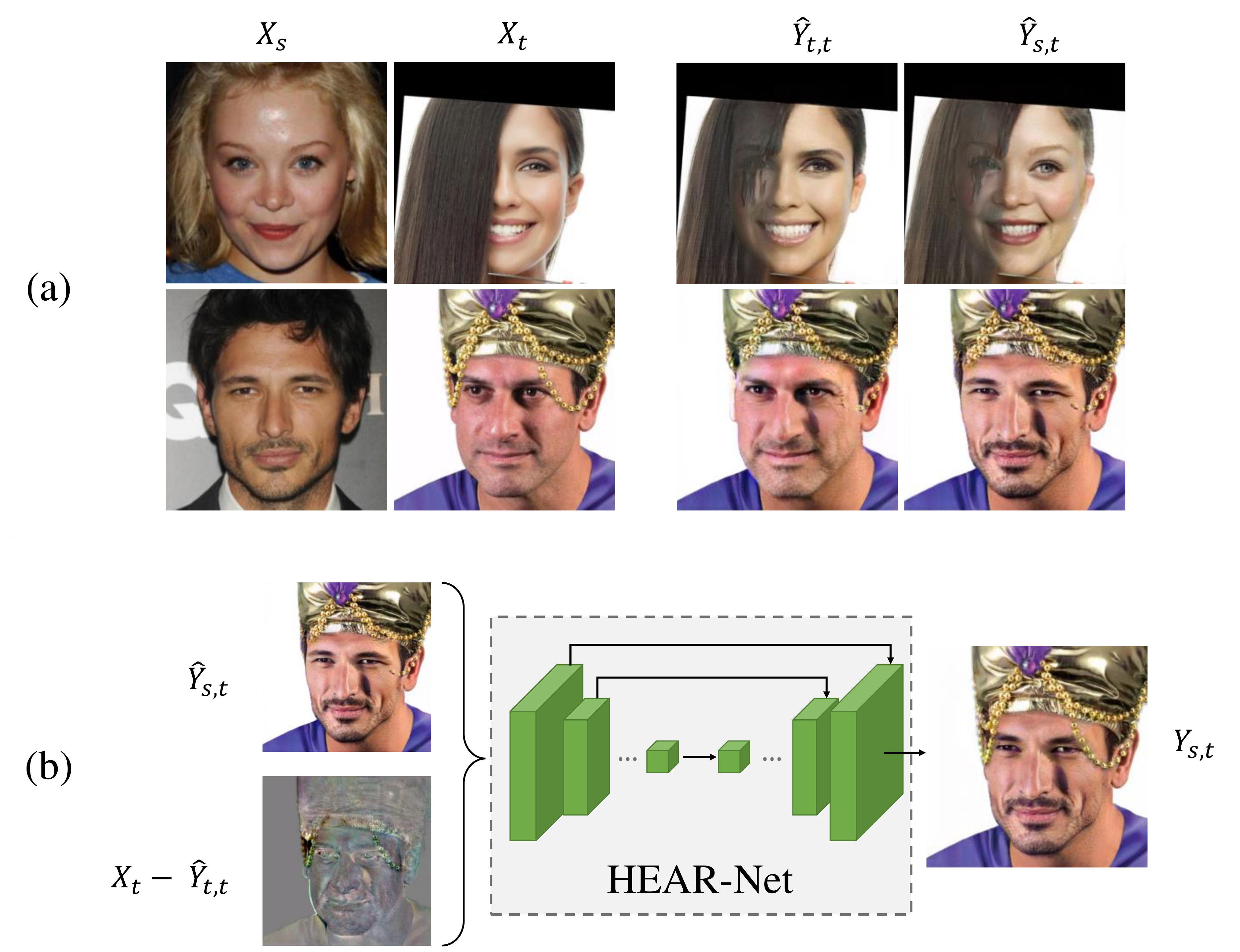}
 \footnotesize
    \caption{HEAR-Net for the second stage. $\hat{Y}_{t,t}$ is the reconstruction of the target image $X_t$, \ie, $\hat{Y}_{t,t}=\texttt{AEI-Net}(X_t, X_t)$. $\hat{Y}_{s,t}$ is the swapped face from the first stage.}
\label{fig:framework_stage2}
\end{figure}

Although the face swap result $\hat{Y}_{s,t}$ generated with AEI-Net in the first stage can well retain target attributes like pose, expression and scene lighting, it often fails to preserve the occlusions appeared on the target face $X_t$.
Previous methods \cite{nirkin2018face,nirkin2019fsgan} address face occlusions with an additional face segmentation network. It is trained on face data containing occlusion-aware face masks, which require lots of manual annotations. Besides, such a supervised approach may hardly recognize unseen occlusion types.

We proposed a heuristic method to handle facial occlusions. As shown in Figure~\ref{fig:framework_stage2}(a), when the target face was occluded, some occlusions might disappear in the swapped face, \eg., the hair covering the face or the chains hang from the turban. Meanwhile, we observe that if we feed the same image as both the source and target images into a well trained AEI-Net, these occlusions would also disappear in the reconstructed image. Thus, the error between the reconstructed image and its input can be leveraged to locate face occlusions. We call it the \emph{heuristic error} of the input image, since it heuristically indicates where anomalies happen.

Inspired by the above observation, we make use of a novel HEAR-Net to generate a refined face image. 
We first get the heuristic error of the target image as
\vspace{-0.5em}
\begin{equation}
\vspace{-0.5em}
\Delta Y_t = X_t - \texttt{AEI-Net}(X_t, X_t).
\end{equation}
Then we feed the heuristic error $\Delta Y_t$ and the result of the first stage $\hat{Y}_{s,t}$ into a U-Net structure, and obtain the refined image $Y_{s,t}$:
\vspace{-0.5em}
\begin{equation}
\vspace{-0.5em}
\label{eqn:Y_s_t}
Y_{s,t} = \texttt{HEAR-Net}(\hat{Y}_{s,t}, \Delta Y_t).
\end{equation}
The pipeline of HEAR-Net is illustrated in Figure \ref{fig:framework_stage2}(b).

We train HEAR-Net in a fully self-supervised way, without using any manual annotations. 
Given any target face image $X_t$, with or without occlusion regions, we utilize the following losses for training HEAR-Net. The first is an identity preservation loss to preserve the identity of the source. Similar as stage one, it is formulated as 
\vspace{-0.5em}
\begin{equation}
\vspace{-0.5em}
\label{eqn:L_id2}
\mathcal{L}'_{id} =  1 - cos(\bm{z}_{id}(Y_{s, t}), \bm{z}_{id}(X_s) ).
\end{equation}
The change loss $\mathcal{L}'_{chg}$ guarantees the consistency between the results of the first stage and the second stage:
\vspace{-0.5em}
\begin{equation}
\vspace{-0.5em}
\label{eqn:L_change}
\mathcal{L}'_{chg} =  \left|\hat{Y}_{s,t} - Y_{s,t}\right|.
\end{equation}
The reconstruction loss $\mathcal{L}'_{rec}$ restricts that the second stage is able to reconstruct the input when the source and target images are the same:

\vspace{-0.5em} 
\begin{equation}
\vspace{-0.5em}
\label{eqn:L_rec}
\mathcal{L}'_{rec} = 
\begin{cases}
 \frac{1}{2} \left\|Y_{s,t} - X_t\right\|_2^2 & \textit{if}~ X_t = X_s \\
0 & \textit{otherwise}
\end{cases}.
\end{equation} 

Since the number of occluded faces is very limited in most face datasets, we propose to augment data with synthetic occlusions.
The occlusions are randomly sampled from a variety of datasets, including the EgoHands \cite{bambach2015lending}, GTEA Hand2K \cite{fathi2011learning,li2015delving,li2013learning} and ShapeNet \cite{chang2015shapenet}.
They are blended onto existing face images after random rotations, rescaling and color matching. Note that \emph{we do not utilize any occlusion mask supervision during training, even from these synthetic occlusions}.

Finally, HEAR-Net is trained with a sum of above losses: 
\vspace{-0.5em}
\begin{equation}
\vspace{-0.5em}
\label{eqn:L_stage2}
\mathcal{L}_\texttt{HEAR-Net} = \mathcal{L}'_{rec} +\mathcal{L}'_{id} + \mathcal{L}'_{chg}.
\end{equation}

\section{Experiments}

\noindent\textbf{Implementation Detail}: 
For each face image, we first align and crop the face using five point landmarks extracted with \cite{chen2014joint}, the cropped image is of size $256\times 256$ covering the whole face, as well as some background regions. The number of attribute embeddings in AEI-Net is set to $n=8$ (Equation~\ref{eqn:attribute_embedding}). The number of downsamples/upsamples in HEAR-Net is set to $5$. Please refer to the supplemental material for more details concerning the network structure and training strategies.

The AEI-Net is trained using CelebA-HQ \cite{karras2017progressive}, FFHQ \cite{karras2019style} and VGGFace \cite{parkhi2015deep}. 
While the HEAR-Net is trained using only a portion of faces that have Top-$10\%$ heuristic errors in these datasets, 
and with additional augmentations of synthetic occlusions. 
Occlusion images are randomly sampled from the EgoHands \cite{bambach2015lending}, GTEA Hand2K \cite{fathi2011learning,li2015delving,li2013learning} and object renderings from ShapeNet \cite{chang2015shapenet}.

\subsection{Comparison with Previous Methods}
\begin{figure}[t]
\centering
 \includegraphics[width=\linewidth,trim={0 8pt 0 0},clip]{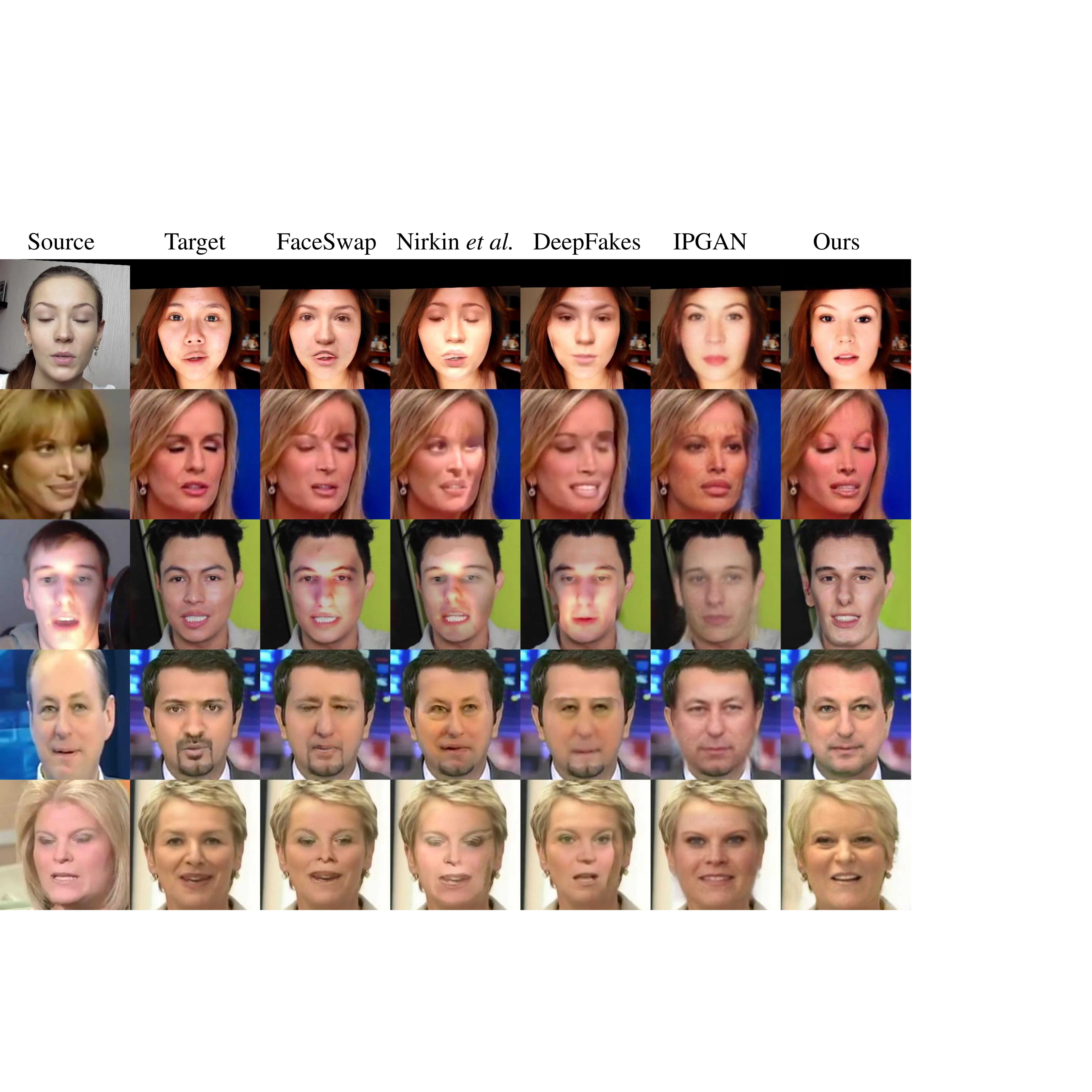}
 \footnotesize
 \caption{Comparison with FaceSwap \cite{faceswap}, Nirkin \etal \cite{nirkin2018face}, DeepFakes \cite{deepfake}, IPGAN \cite{Bao_ipgan} on FaceForensics++ \cite{rossler2019faceforensics++} face images. Our results better preserve the face shapes of the source identities, and are also more faithful to the target attributes (\eg lightings, image resolutions).}
\label{fig:compare_all_no_fsgan}
\vspace{-1em}
\end{figure}

\begin{figure}[t]
\centering
 \includegraphics[width=\linewidth]{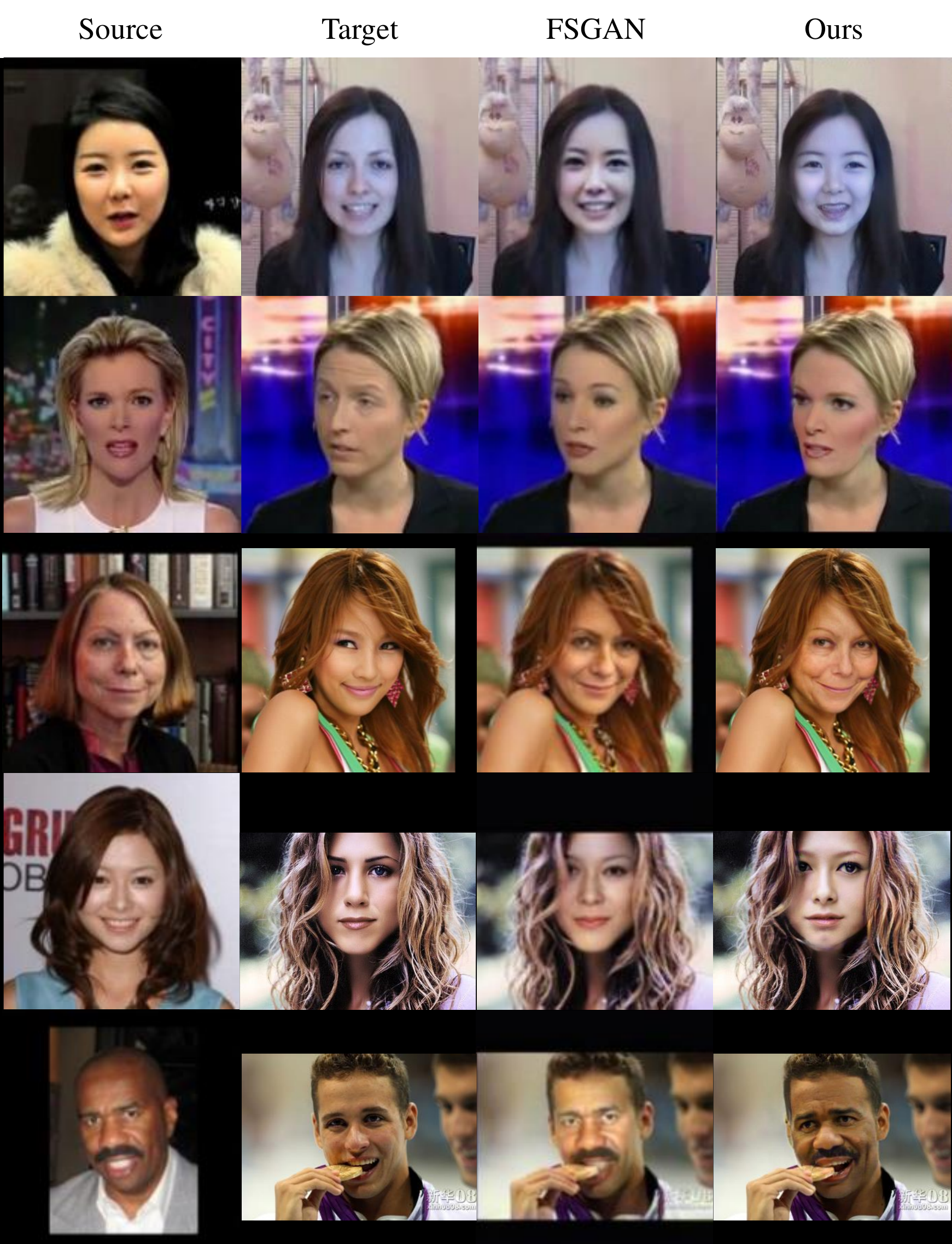}
 \footnotesize
 \caption{Comparison with FSGAN \cite{nirkin2019fsgan}. 
Besides the advantages in face quality and fidelity to inputs, our results preserve common occlusions as good as FSGAN. 
Please also refer to Figures \ref{fig:teaser}, \ref{fig:effect_of_hear} and \ref{fig:wild} for more challenging cases.}
\label{fig:compare_fsgan}
\vspace{-1em}
\end{figure}

\noindent\textbf{Qualitative Comparison}:
We compare our method with FaceSwap \cite{faceswap}, Nirkin \etal \cite{nirkin2018face}, DeepFakes \cite{deepfake} and IPGAN \cite{Bao_ipgan} on the FaceForensics++ \cite{rossler2019faceforensics++} test images in Figure \ref{fig:compare_all_no_fsgan}. 
Comparison with the latest work FSGAN \cite{nirkin2019fsgan} is shown in Figure \ref{fig:compare_fsgan}.
We can see that, since FaceSwap, Nirkin \etal, DeepFakes, and FSGAN all follow the strategy that first synthesizing the inner face region then blending it into the target face, as expected, they suffer from the blending inconsistency. 
All faces generated by these methods share exactly the same face contours with their target faces, and ignore the source face shapes (Figure \ref{fig:compare_all_no_fsgan} rows 1-4, Figure \ref{fig:compare_fsgan} rows 1-2). Besides, their results can not well respect critical informations from the target image, such as the lighting (Figure \ref{fig:compare_all_no_fsgan} row 3, Figure \ref{fig:compare_fsgan} rows 3-5), the image resolutions (Figure \ref{fig:compare_all_no_fsgan} rows 2 and 4). IPGAN \cite{Bao_ipgan} suffers from decreased resolutions in all samples, due to its single-level attributes representation. IPGAN cannot well preserve expression of the target face, such as the closed eyes (Figure \ref{fig:compare_all_no_fsgan} row 2).

Our method addresses all these issues well. We achieve higher fidelity by well preserving the face shapes of the source (instead of the target), and faithfully respecting the lighting and image resolution of the target (instead of the source). Our method also has the ability to go beyond FSGAN~\cite{nirkin2019fsgan} to handle occlusions.

\noindent\textbf{Quantitative Comparison}:
The experiment is constructed on FaceForensics++ \cite{rossler2019faceforensics++} dataset. For FaceSwap \cite{faceswap} and DeepFakes \cite{deepfake}, the test set consists of 10K face images for each method by evenly sampled 10 frames from each video clip. For IPGAN~\cite{Bao_ipgan}, Nirkin \etal \cite{nirkin2018face} and our method, $10K$ face images are generated with the same source and target image pairs as the other methods.
Then we conduct quantitative comparison with respect to three metrics: \emph{ID retrieval}, \emph{pose error} and \emph{expression error}.

\begin{table}[t]
\begin{center}
\small
\begin{tabular}{c|ccc}
method & ID retrieval $\uparrow$ & pose $\downarrow$ & expression $\downarrow$  \\
\hline
DeepFakes \cite{deepfake} & 81.96  & 4.14 &2.57\\  
FaceSwap \cite{faceswap}  & 54.19  & \textbf{2.51} & 2.14\\ 
Nirkin \etal \cite{nirkin2018face} & 76.57 & 3.29& 2.33\\ 
IPGAN \cite{Bao_ipgan} & 82.41 & 4.04 & 2.50 \\ 
\hline
Ours & \textbf{97.38} & 2.96 & \textbf{2.06} \\ 
\end{tabular}
\caption{Comparison on FaceForensics++ videos.}
\label{table:ff++}
\end{center}
\vspace{-2em}
\end{table}

We extract identity vector using a different face recognition model \cite{wang2018cosface} and adopt the cosine similarity to measure the identity distance.
For each swapped face from the test set, we search the nearest face in all FaceForensics++ original video frames and check whether it belongs to the correct source video.
The averaged accuracy of all such retrievals is reported as the \emph{ID retrieval} in Table \ref{table:ff++}, serving to measure identity preservation ability. 
Our method achieves higher \emph{ID retrieval} score with a large margin.

We use a pose estimator \cite{ruiz2018fine} to estimate head pose and a 3D face model \cite{chaudhuri2019joint} to retrieve expression vectors. We report the $\mathcal{L}$-2 distances of pose and expression vectors between the swapped face and its target face in Table \ref{table:ff++} as the \emph{pose} and the \emph{expression} errors. Our method is advantageous in expression preservation while comparable with others in pose preservation. We do not use the face landmark comparison as~\cite{nirkin2019fsgan}, since face landmarks involve identity information which should be inconsistent between the swapped face and the target face.

\noindent\textbf{Human Evaluation}: Three user studies are conducted to evaluate the performance of the proposed model. We let the users select: i) \emph{the one having the most similar identity with the source face}; ii) \emph{the one sharing the most similar head pose, face expression and scene lighting with the target image}; iii) \emph{the most realistic one}. 
In each study unit, two real face images, the source and the target, and four reshuffled face swapping results generated by FaceSwap \cite{faceswap}, Nirkin \etal \cite{nirkin2018face}, DeepFakes \cite{deepfake} and ours, are presented. We ask users to select one face that best matches our description. 

\begin{figure}[t]
	\centering
	\includegraphics[width=\linewidth]{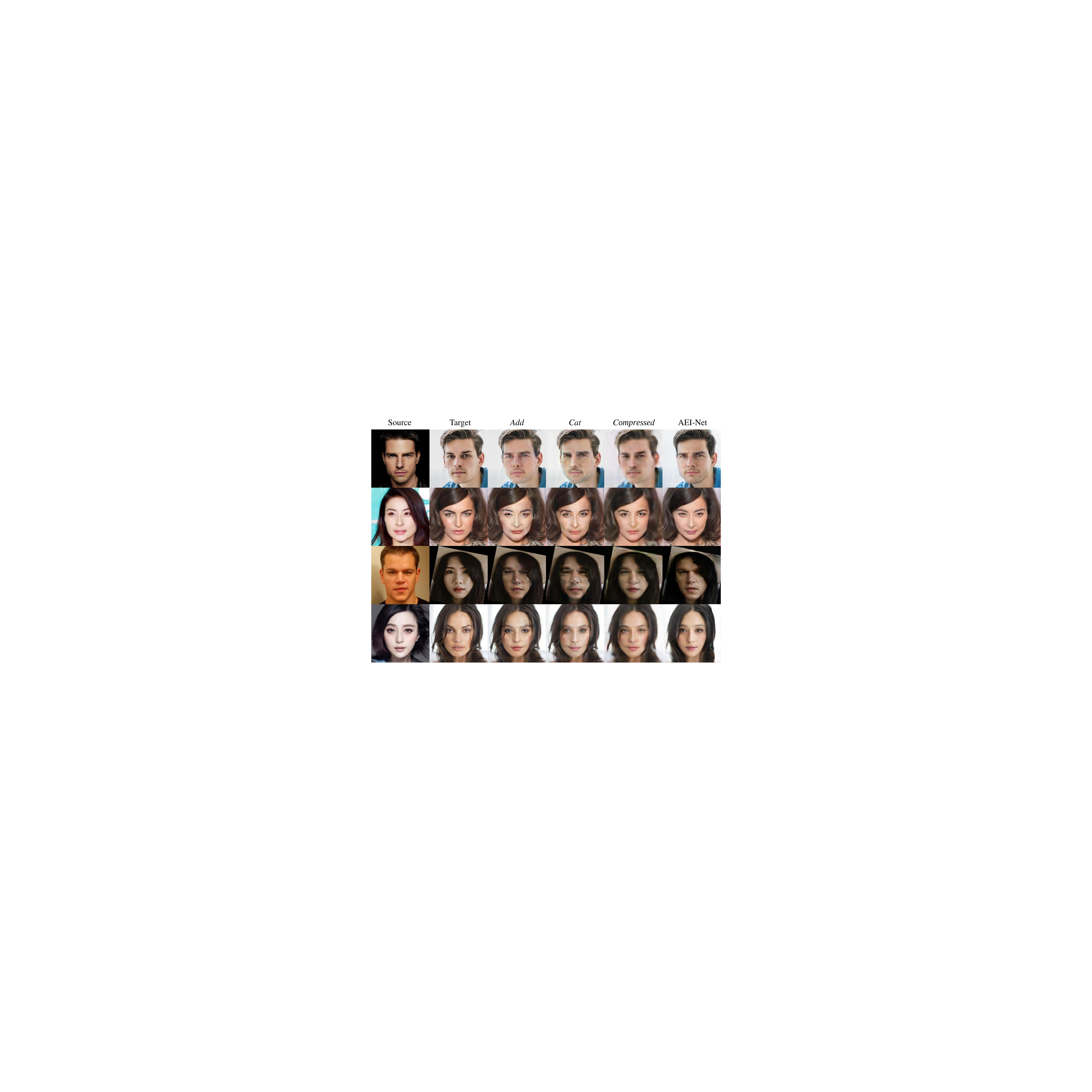}
	\footnotesize
	\caption{Comparing AEI-Net with three baseline models. The two models \emph{Add} and \emph{Cat} are for ablation studies of the adaptive embedding integration. The model \emph{Compressed} is for ablating multi-level attributes representation.}
	\label{fig:ablation_aei}
\end{figure}

\begin{table}[t]
\begin{center}
\small
\begin{tabular}{c|cccc}
method &  id. & attr. & realism  \\
\hline
DeepFakes \cite{deepfake} & 13.7 & 6.8 & 6.1 \\  
FaceSwap \cite{faceswap}  & 12.1 & 23.7 & 6.8 \\ 
Nirkin \etal \cite{nirkin2018face} & 21.3 & 7.4 & 4.2\\ 
Ours & \textbf{52.9} & \textbf{62.1} & \textbf{82.9} \\ 
\end{tabular}
\caption{User study results. We show the averaged selection percentages of each method.}
\label{table:user_study}
\end{center}
\vspace{-2em}
\end{table}

For each user, 20 face pairs are randomly drawn from the 1K FaceForensics++ test set without duplication.
Finally, we collect answers from 100 human evaluators. The averaged selection percentage for each method on each study is presented in Table \ref{table:user_study}.
It shows that our model surpasses the other three methods all in large margins.

\subsection{Analysis of the Framework}
\label{ssec:analysis}

\noindent\textbf{Adaptive Embedding Integration}: 
To verify the necessity of adaptive integration using attentional masks, we compare AEI-Net with two baseline models:
i) \emph{Add}: element-wise plus operations is adopted in AAD layers instead of using masks $\bm{M}^k$ as in Equation \ref{eqn:Mask}. The output activation $\bm{h}_{out}^k$ of this model is directly calculated with $\bm{h}_{out}^k = \bm{A}^k + \bm{I}^k$;
ii) \emph{Cat}: element-wise concatenation is adopted without using masks $\bm{M}^k$. The output activation becomes $\bm{h}_{out}^k = \texttt{Concat}[\bm{A}^k, \bm{I}^k]$. Results of the two baseline models, as well as the AEI-Net, are compared in Figure \ref{fig:ablation_aei}. 
Without a soft mask for fusing embeddings adaptively, the faces generated by baseline models are relatively blurry and contain lots of ghosting artifacts.

\begin{figure}[t]
\centering
 \includegraphics[width=\linewidth]{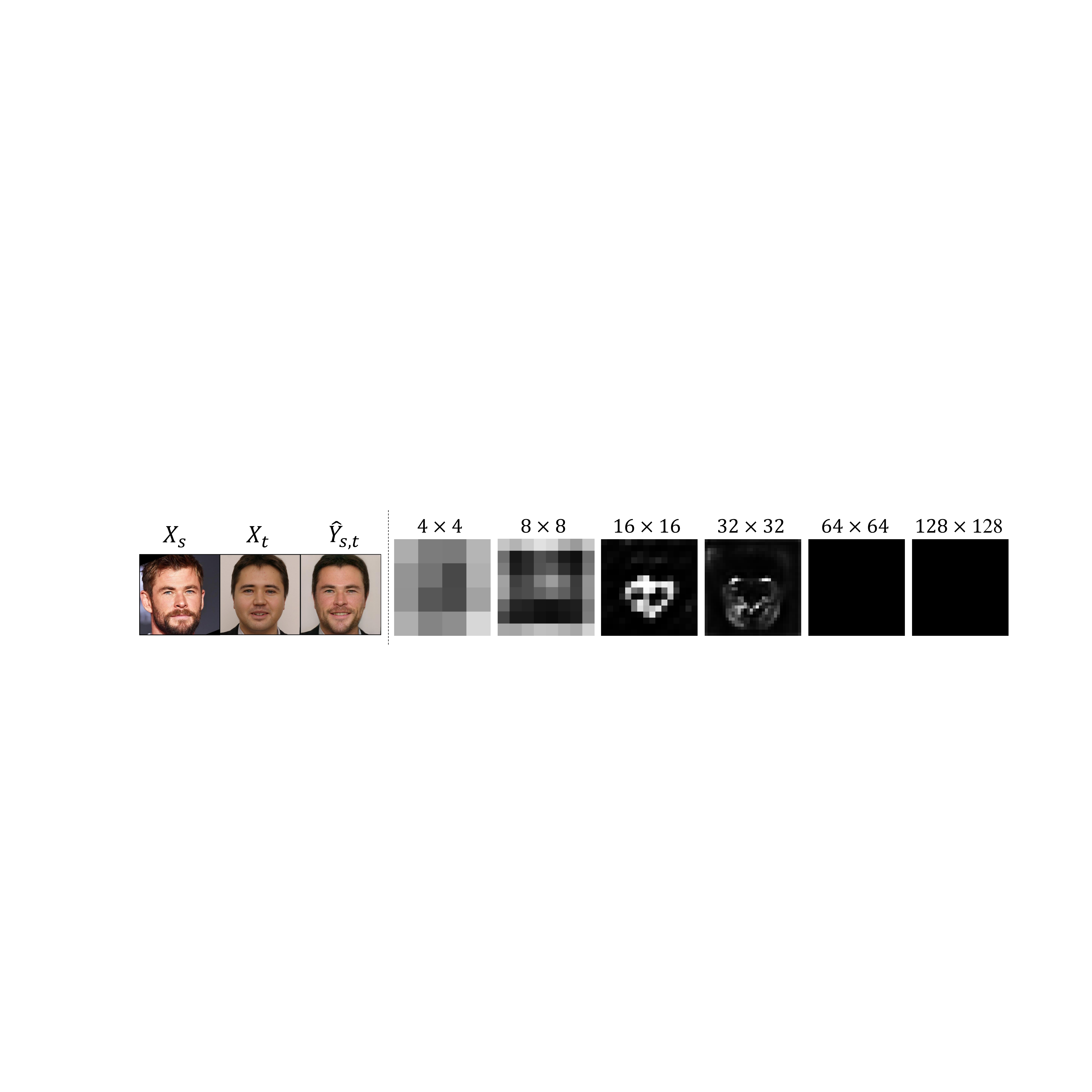}
    \caption{Visualizing attentional masks $\bm{M}^k$ of AAD layers on different feature levels. These visualizations reflect that identity embeddings are mostly effective in low and middle feature levels.}
\label{fig:visualize_aad_masks}
\end{figure}

We also visualize the masks $\bm{M}^k$ of AAD layers on different levels in Figure \ref{fig:visualize_aad_masks}, where a brighter pixel indicates a higher weight for identity embedding in Equation \ref{eqn:Mask}. It shows that the identity embedding takes more effect in low level layers. Its effective region becomes sparser in middle levels, where it activates only in some key regions that strongly relates to the face identity, such as the locations of eyes, mouth and face contours.

\begin{figure}[t]
\centering
 \includegraphics[width=\linewidth]{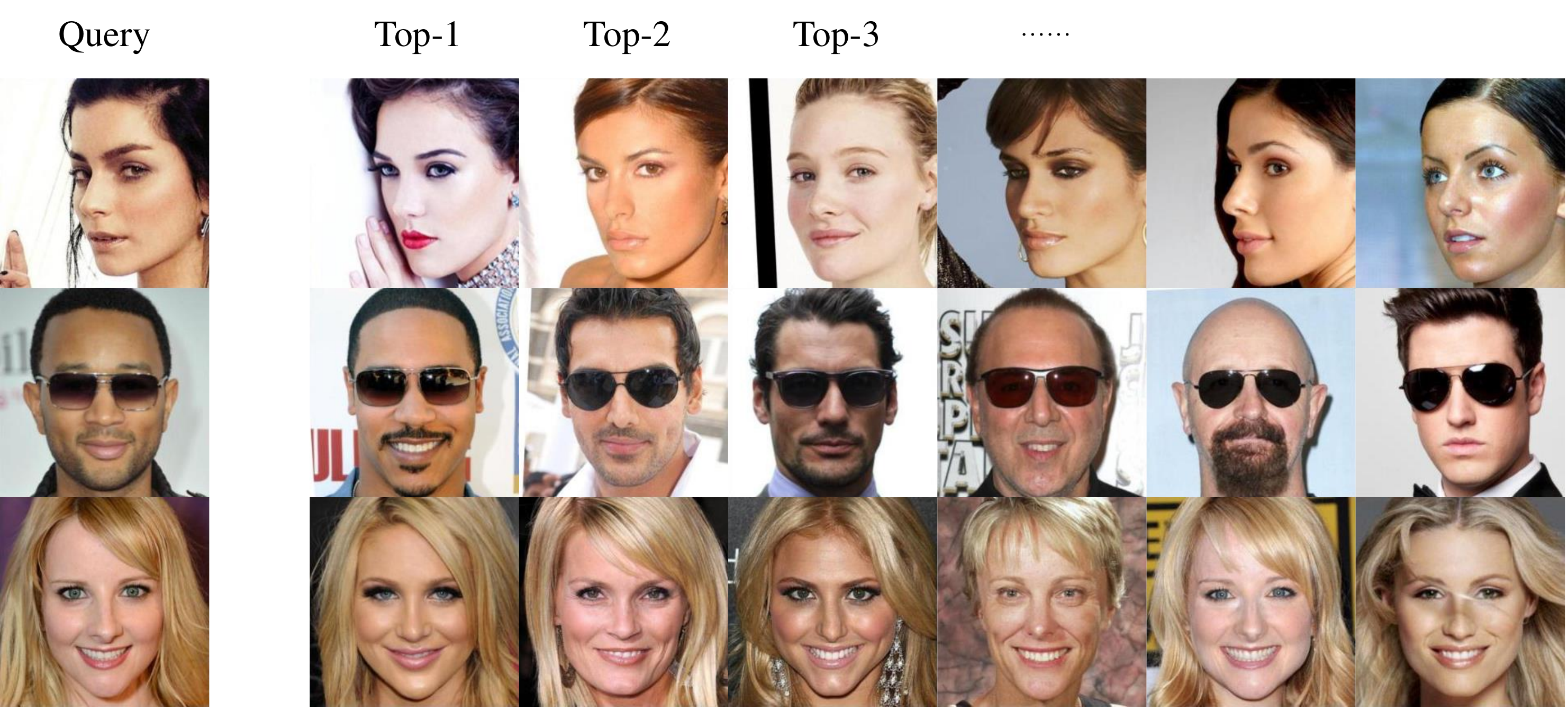}
 \footnotesize
    \caption{Query results using attributes embedding.}
\label{fig:att_query}
\vspace{-0.5em}
\end{figure}

\begin{figure*}[t]
\centering
\includegraphics[width=\linewidth,trim={0 4pt 0 0},clip]{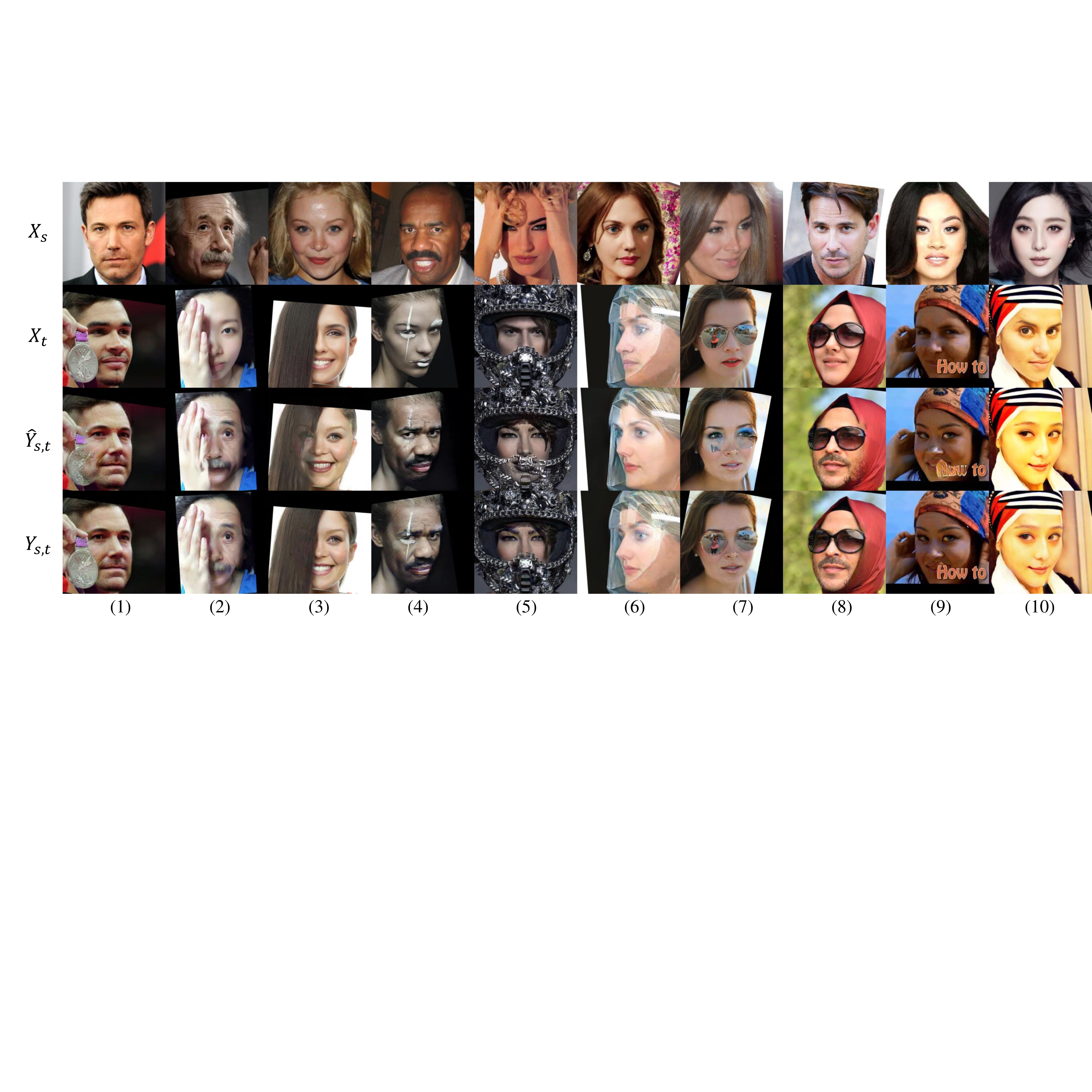}
\caption{Second-stage refining results presenting the strong adaptability of HEAR-Net on various kinds of errors, including occlusions, reflections, slightly shifted pose and color etc.}
\label{fig:effect_of_hear}
\end{figure*}

\begin{figure*}[t]
\centering
 \includegraphics[width=\linewidth]{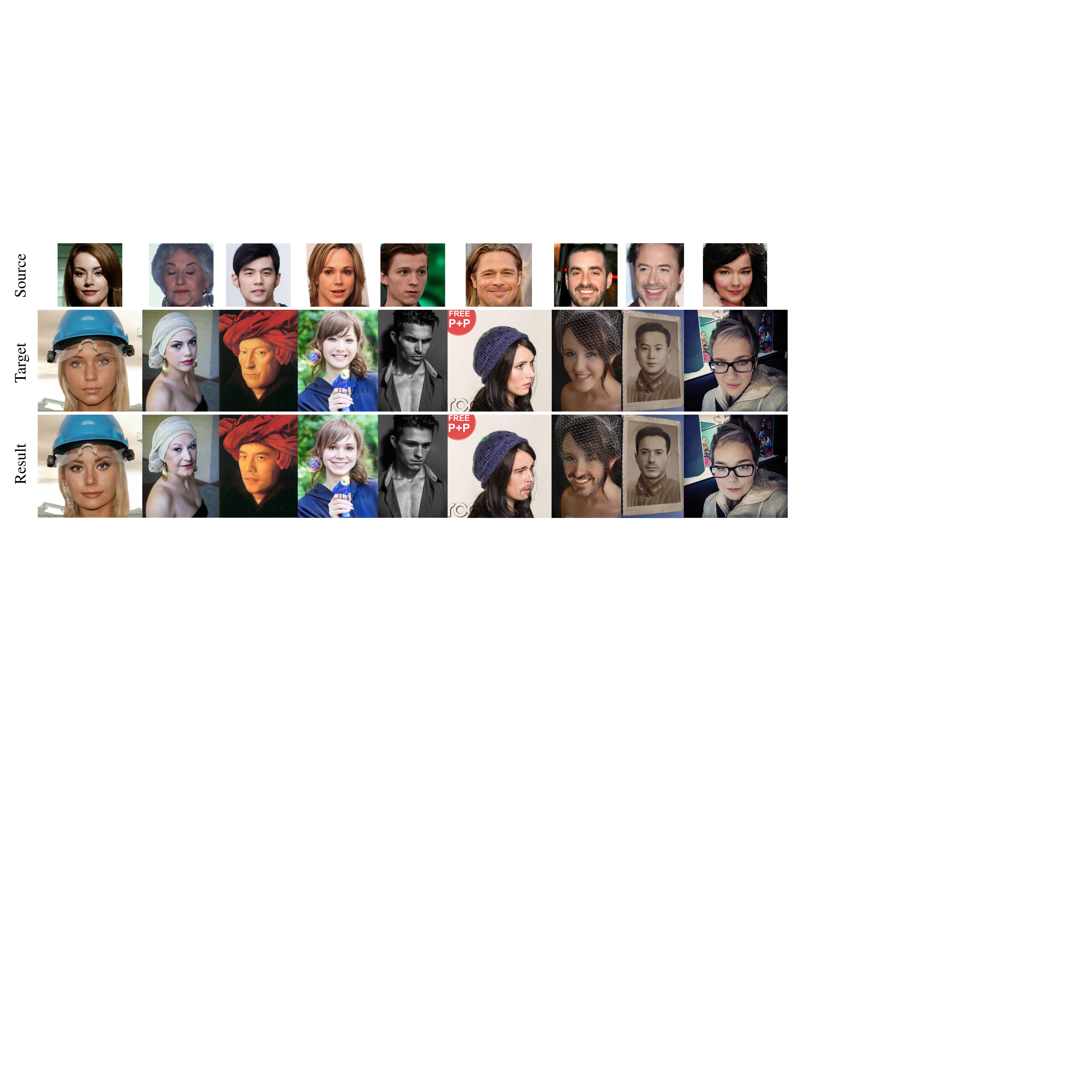}
    \caption{Our face swapping results on wild face images under various challenging conditions. All results are generated using a single well-trained two-stage model.}
\label{fig:wild}
\vspace{-0.5em}
\end{figure*}

\noindent\textbf{Multi-level Attributes}:
To verify whether it is necessary to extract multi-level attributes, we compare with another baseline model called \emph{Compressed}, which shares the same network structure with AEI-Net, 
but only utilizes the first three level embeddings $\bm{z}_{att}^k, k=1,2,3$. Its last embedding $\bm{z}_{att}^3$ is fed into all higher level AAD integrations.  
Its results are also compared in Figure \ref{fig:ablation_aei}. Similar to IPGAN \cite{Bao_ipgan}, its results suffer from artifacts like blurriness, since a lot of attributes information from the target images are lost.

To understand what is encoded in the attributes embedding, we concatenate the embeddings $\bm{z}_{att}^k$ (bilinearly upsampled to $256\times 256$ and vectorized) from all levels as a unified attribute representation. We conduct PCA to reduce vector dimensions as $512$. 
We then perform tests querying faces from the training set with the nearest $\mathcal{L}$-2 distances of such vectors. The three results illustrated in Figure \ref{fig:att_query} verify our intention, that the attributes embeddings can well reflect face attributes, such as the head pose, hair color, expression and even the existence of sunglasses on the face. Thus it also explains why our AEI-Net sometimes can preserve occlusions like sunglasses on the target face even without a second stage (Figure \ref{fig:effect_of_hear}(8)).

\noindent\textbf{Second Stage Refinement}:
Multiple samples are displayed with both one-stage results $\hat{Y}_{s,t}$ and two-stage results $Y_{s,t}$ in Figure \ref{fig:effect_of_hear}. 
It shows that the AEI-Net is able to generate high-fidelity face swapping results, but sometimes its output $\hat{Y}_{s,t}$ does not preserve occlusions in the target.
Fortunately, the HEAR-Net in the second stage is able to recover them. 

The HEAR-Net can handle occlusions of various kinds, such as the medal (1), hand (2), hair (3), face painting (4), mask (5), translucent object (6), eyeglasses (7), headscarf (8) and floating text (9). 
Besides, it is also able to correct the color-shift that might occasionally happen in $\hat{Y}_{s,t}$ (10). 
Moreover, the HEAR-Net can help rectify the face shape when the target face has a very large pose (6).

\subsection{More Results on Wild Faces}

Finally, we demonstrate the strong capability of FaceShifter by testing on wild face images downloaded from Internet. 
As shown in Figure \ref{fig:wild}, our method can handle face images under various conditions, including large poses, uncommon lightings and occlusions of very challenging kinds.

\section{Conclusions}

In this paper, we proposed a novel framework named FaceShifter for high fidelity and occlusion aware face swapping. The AEI-Net in the first stage adaptively integrates the identity and the attributes for synthesizing high fidelity results. The HEAR-Net in the second stage recovers anomaly region in a self-supervised way without any manual annotations.  
The proposed framework shows superior performance in generating realistic face images given any face pairs without subject specific training. 
Extensive experiments demonstrate that the proposed framework significantly outperforms previous face swapping methods.


{\small
\bibliographystyle{ieee_fullname}
\bibliography{egbib}
}

\cleardoublepage
\appendix

\section{Network Structures}

\begin{figure}[b]
\centering
\includegraphics[width=0.9\linewidth]{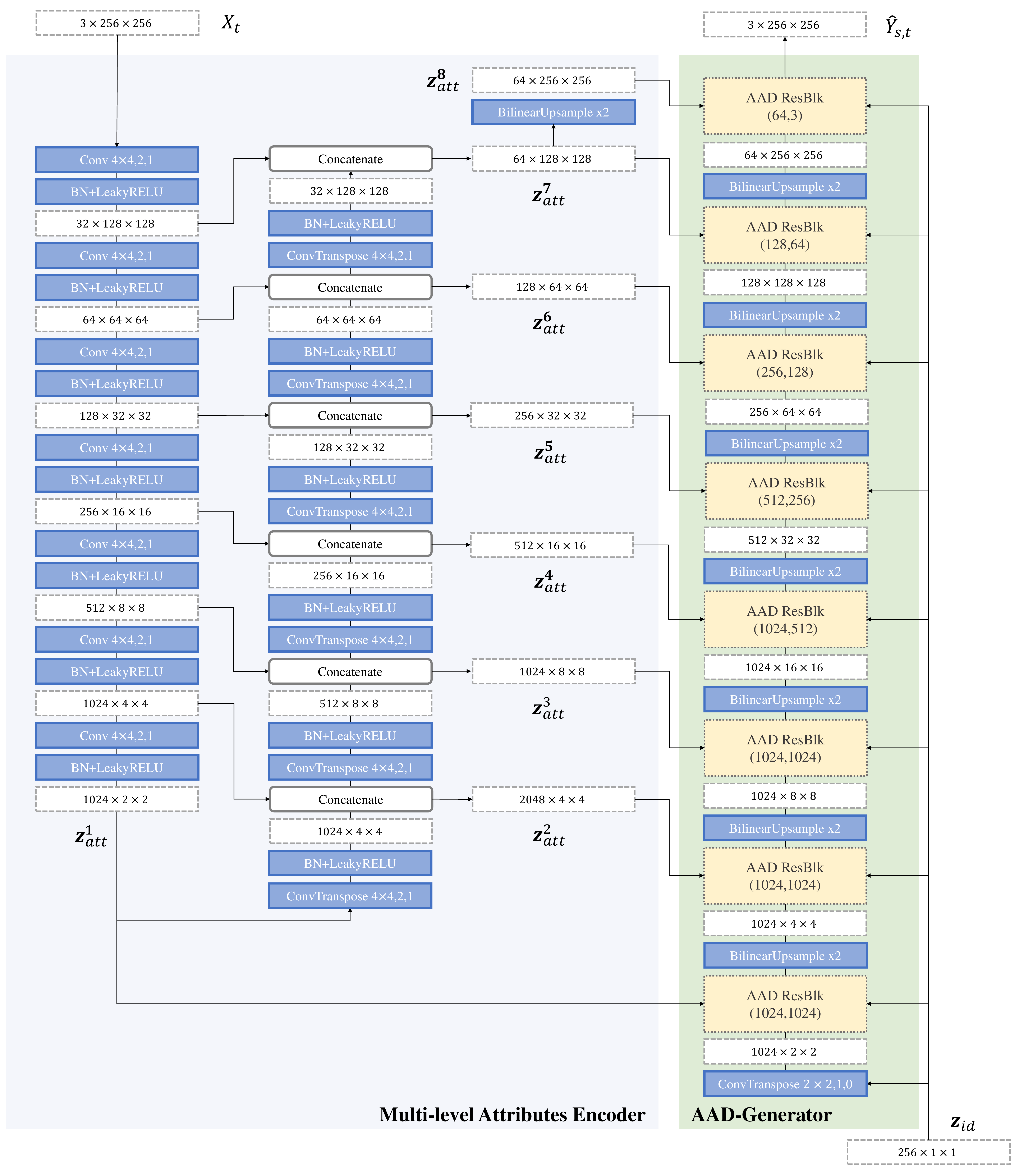}
\includegraphics[width=0.8\linewidth]{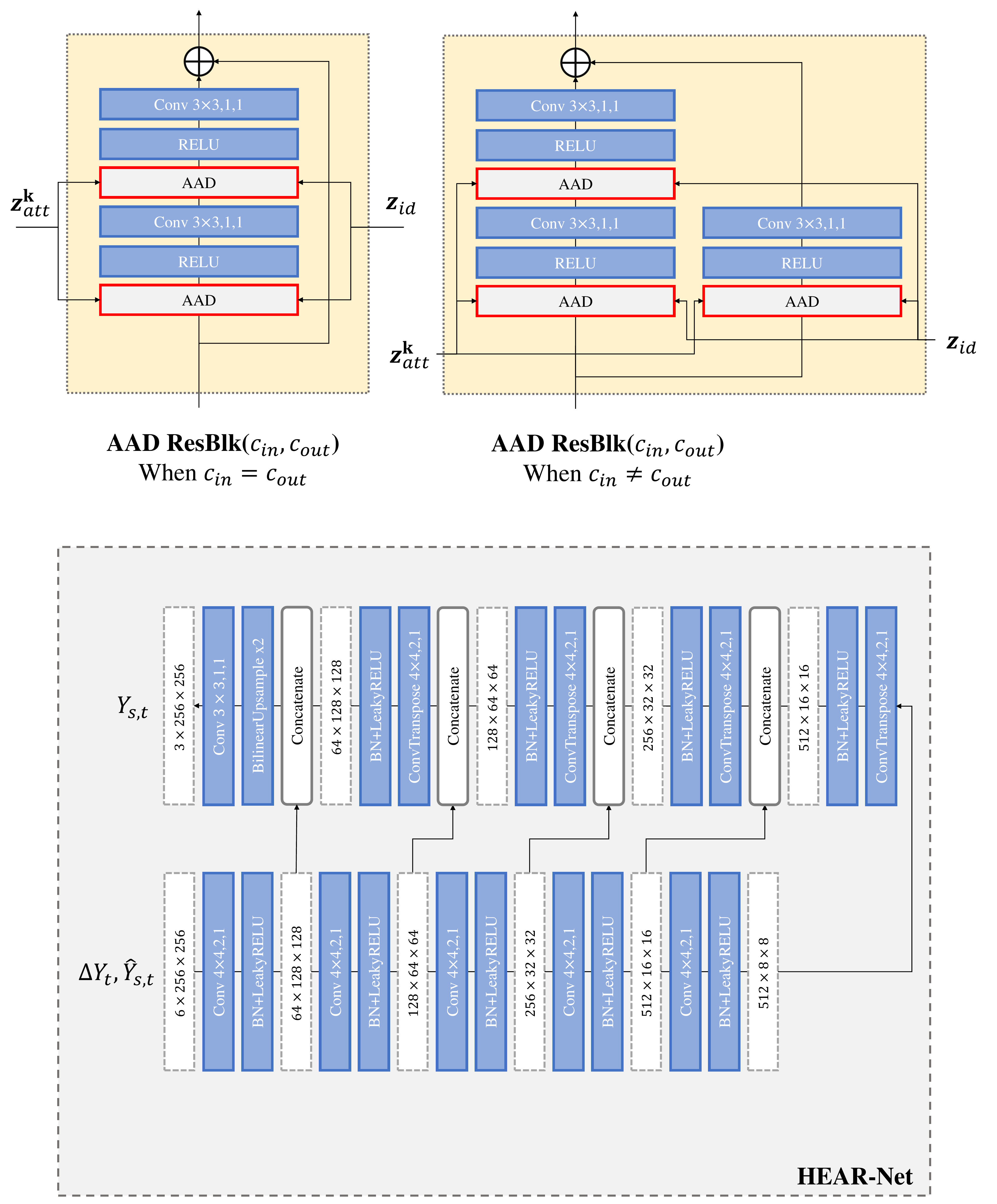}
\caption{Network structures. \emph{Conv k,s,p} represents a Convolutional Layer with kernel size $k$, stride $s$ and padding $p$. \emph{ConvTranspose k,s,p} represents a Transposed Convolutional Layer with kernel size $k$, stride $s$ and padding $p$. All \emph{LeakyReLU}s have $\alpha=0.1$. \emph{AAD ResBlk($c_{in}$, $c_{out}$)} represents an AAD ResBlk with input and output channels of $c_{in}$ and $c_{out}$.}
\label{fig:network}
\end{figure}

Detailed structures of the AEI-Net and the HEAR-Net are given in Figure \ref{fig:network}.

\section{Training Strategies}

In AEI-Net, we use the same multi-scale discriminator as \cite{park2019semantic} with the adversarial loss implemented as a hinge loss. 
The ratio of training samples that have $X_t!=X_s$ is $80\%$ when training the AEI-Net, it is $50\%$ when training the HEAR-Net.
ADAM \cite{kingma2014adam} with $\beta_1=0, \beta_2=0.999, lr=0.0004$ is used for training all networks. The AEI-Net is trained with 500K steps while the HEAR-Net is trained with 50K steps, both using 4 P40 GPUs with 8 images per GPU. We use synchronized mean and variance computation, i.e., these statistics are collected from all the GPUs.

\begin{figure}[h]
\centering
\includegraphics[width=\linewidth]{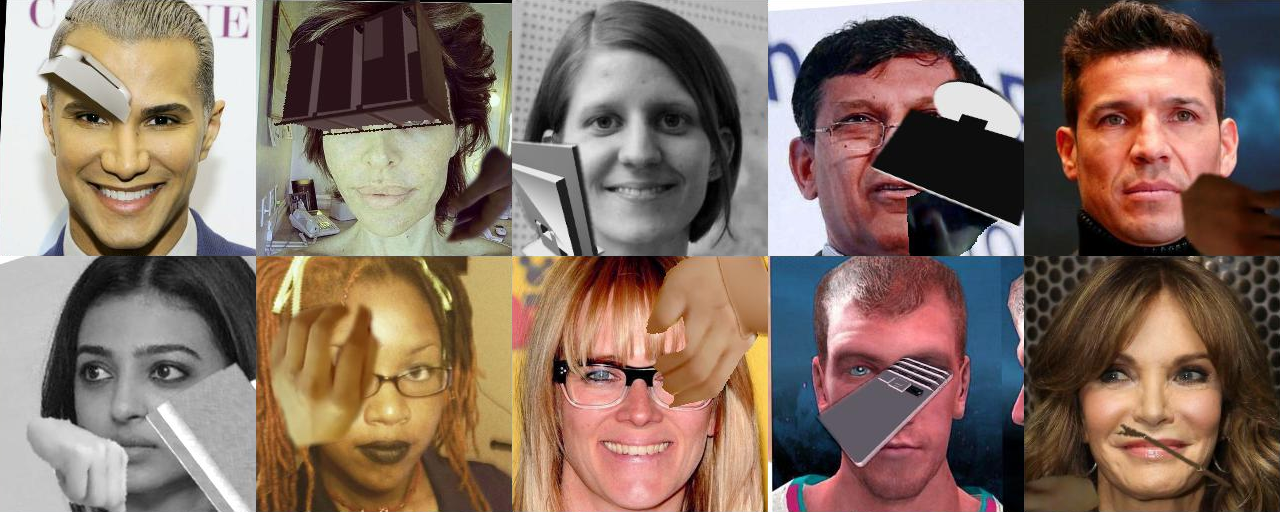}
\caption{Augmentation with synthetic occlusions.}
\label{fig:augmentation}
\end{figure}

Besides sampling hand images from the EgoHands \cite{bambach2015lending} and GTEA Hand2K \cite{fathi2011learning,li2015delving,li2013learning}, we use a public code\footnote{https://github.com/panmari/stanford-shapenet-renderer} for rendering ShapeNet \cite{chang2015shapenet} objects in occlusion data augmentation. Some synthetic occlusions are shown in Figure \ref{fig:augmentation}.

\end{document}